\documentclass[runningheads]{llncs}
\usepackage{graphicx}
\usepackage{comment}
\usepackage{multirow}
\begin{document}
%

\title{Evaluation of RGB-D SLAM in Large Indoor Environments}
%
%
\author{Kirill Murayvev\inst{}\orcidID{0000-0001-5897-0702} \and
Konstantin Yakovlev\inst{}\orcidID{0000-0002-4377-321X}}
\authorrunning{K. Muravyev and K. Yakovlev}
%
\institute{Federal Research Center ``Computer Science and Control'' of Russian Academy of Sciences, Moscow, Russia
\\
\email{muraviev@isa.ru, yakovlev@isa.ru}
}
\maketitle              
\begin{abstract}


\newcommand\blfootnote[1]{%
  \begingroup
  \renewcommand\thefootnote{}\footnote{#1}%
  \addtocounter{footnote}{-1}%
  \endgroup
}

\blfootnote{This is a pre-print of the paper accepted to ICR 2022 conference}

Simultaneous localization and mapping (SLAM) is one of the key components of a control system that aims to ensure autonomous navigation of a mobile robot in unknown environments. In a variety of practical cases a robot might need to travel long distances in order to accomplish its mission. This requires long-term work of SLAM methods and building large maps. Consequently the computational burden (including high memory consumption for map storage) becomes a bottleneck. Indeed, state-of-the-art SLAM algorithms include specific techniques and optimizations to tackle this challenge, still their performance in long-term scenarios needs proper assessment. To this end, we perform an empirical evaluation of two widespread state-of-the-art RGB-D SLAM methods, suitable for long-term navigation, i.e. RTAB-Map and Voxgraph. We evaluate them in a large simulated indoor environment, consisting of corridors and halls, while varying the odometer noise for a more realistic setup. We provide both qualitative and quantitative analysis of both methods uncovering their strengths and weaknesses. We find that both methods build a high-quality map with low odometry noise but tend to fail with high odometry noise. Voxgraph has lower relative trajectory estimation error and memory consumption than RTAB-Map, while its absolute error is higher.

\end{abstract}

\section{Introduction}

Making a robotic systems fully autonomous is an important problem for modern researchers~\cite{gonzalez2017supervisory,papachristos2019autonomous,tang2019topological,choi2020development}. In order to operate autonomously, the system needs to know it's position on the map (environment). In case of indoor or underground environment, global position estimation like GPS is unable to work, so a robotic system has to estimate its position using its own sensors.  If the environment is unknown, the map is also need to be built. So, for solving this problem, Simultaneous Localization and Mapping (SLAM) techniques are need.

Traditional SLAM methods like ORB-SLAM \cite{mur2015orb} build a metric map  of the environment matching features extracted from input images or laser scans. They map only feature cooridnates, leading to building a sparse map. Path planning in such map is difficult because a planner may create path between feature points through unmapped obstacles.

On the other hand, direct SLAM methods like LSD-SLAM \cite{engel2014lsd} and learning-based methods like DROID-SLAM \cite{teed2021droid} build dense map but require high computational costs to update the map, especially in large environments. So, using them in a real robotic system for long-term navigation is highly restricted.

A possible solution for dense mapping a large environment is utilizing stereo or depth information in feature-based methods and its reprojecting during SLAM process. Such technique is implemented in RTAB-Map method \cite{labbe2019rtab}. RTAB-Map builds a graph of key frames and involves a comprehensive memory management approach which is detailly described in section \ref{sec:rtabmap}. The key frames are stored at three levels of memory that helps maintain large maps and effectively close loops.

Another tecnhique of effective long-lasting SLAM and odometry error elimination is building a graph of small submaps instead of keyframes. A submap is a map of a short segment of the environment built by a SLAM algorithm. It has an anchor pose (e.g. pose of the robot at the time of entering the submap, or pose of the submap's centroid), and these poses are structured in a graph. Pose graph and map correction could be done using graph optimization and submap fusing techniques. Such strategy is implemented in Voxgraph \cite{reijgwart2019voxgraph} algorithm.
In this work we do empirical evaluation of Voxgraph and  RTAB-MAP methods. We conduct a series of experiments in photo-realistic Habitat simulator \cite{savva2019habitat} in our university building model with 100m corridor, halls and rooms. We evaluate absolute and relative trajectory errors for both methods over trajectories of length up to 1km. We provide detailed experimental analysis in section \ref{sec:experiments}.


\section{Related Work}



\subsection{SLAM algorithms}

Simultaneous localization and mapping has long history of study and is widely researched. First SLAM methods \cite{smith1988stochastic} use extended Kalman filter (EKF) to estimate robot's trajectory from noised sensor data. More recent works \cite{klein2007parallel} \cite{mur2015orb} extract features from input images and track robot motion matching these features. Also these methods use global optimization techniques like Bundle Adjustment \cite{triggs1999bundle} to perform loop closure (global trajectory and map optimization in case of re-visiting a previously visited place). ORB-SLAM2 method \cite{mur2017orb} includes feature extraction, tracking, local mapping, and loop closing parts, and achieves trajectory estimation error of approximately 1\% on KITTI dataset \cite{geiger2013vision}. However, this method builds sparse map (only feature points are mapped), which is not suitable for path planning.

Besides traditional feature-based SLAM methods, deep learning-based methods are also actively developing. In method \cite{min2021voldor+} neural network is used for optical flow estimation in order to accurately estimate robot's position. In work \cite{teed2021droid} SLAM is performed by a fully learning-based pipeline, including feature extraction and bundle adjustment. Both these methods build dense map and have small error, however, they require powerful GPU to operate, so use of these methods in a mobile robotic system is difficult.

One of the most common classical SLAM methods is RTAB-Map \cite{labbe2019rtab}. It tracks robot motion and builds the map matching features extracted from images. It builds dense map using depth information from RGB-D, stereo or lidar data. The map is stored in 3D point cloud format and projected into 2D occupancy grid which is convenient for path planning. Also RTAB-Map utilizes a complex memory management strategy which helps make effective mapping and loop closing even in large areas. We use this method in our comparison.

For navigation in large environments, map decomposition into submaps is also used. Such decomposition helps effectively optimize map and trajectory, significantly reduce path planning costs, and eliminate odometry error accumulation. In recent work \cite{gomez2020hybrid} the map is represented as a graph of submaps, one submap for each room or hallway. For navigation, a graph of doors and passages is used. This approach reduces planning time in two orders of magnitude (0.5s vs 67s) in comparison with global metric map. However, despite planning effectiveness, this method requires high computational resources for doors detection to divide the map into submaps.

Another submap-based approach is Voxgraph \cite{reijgwart2019voxgraph}. Submaps are switched in a fixed period of time. For submap creation, Voxblox \cite{oleynikova2017voxblox} method is used. Loop closure is performed in two levels: local (in Voxblox, using ICP \cite{besl1992method} method), and global, using graph optimization methods. In work \cite{schmid2021unified}, Voxgraph managed to build proper map during exploration of a large indoor area with severe odometry drift. We used Voxgraph method in our experimental evaluation.

\subsection{Navigation and mapping in large areas}

Navigation in large environments and mapping large areas is an important problem for industrial and service robots. There are some works for building a map up to city-scale size \cite{niijima2020city}. In work \cite{niijima2020city}, pre-built pointcloud maps and cloud global map data like OpenStreetMap are used. In work \cite{gomez2020hybrid}, a hybrid metric-topological approach is used to map large areas. With this approach, a robot successfully explored an indoor area of 1137 $m^2$ with 3x memory reduction comparing with a metric SLAM approach. 

Another metric-topological approach for large area navigation is proposed in work \cite{schmuck2016hybrid}. In that work, a global map is divided into submaps - cubes of fixed size. Each submap is built by a conventional feature-based SLAM. During loop closure, only submaps with significant changes are modified, the other submaps only change their position. Indoor and outdoor tests on a robot with 600m and 1200m trajectories show that the method is able to maintain and quickly update large maps.

\subsection{Long-term SLAM evaluation}

SLAM quality estimation has prolonged history of study. There are a lot of benchmarks for SLAM evaluation, like TUM \cite{sturm12iros}, EuRoC \cite{burri2016euroc}, KITTI \cite{geiger2013vision}. With these benchmarks, a lot of comparative analysis of SLAM systems is done \cite{bokovoy2021assessment} \cite{mingachev2021comparative} \cite{zhang2021survey}. However, most of these works evaluate SLAM algorithms in short trajectories and small environments. There are some bencmarks for long-term SLAM evaluation like OpenLORIS \cite{shi2020we} and SLAM robustness evaluation research like \cite{hong2021radar}. These works examine mostly SLAM robustness to environmental conditions changes (like illumination, weather, etc). In our work, we evaluate SLAM memory and computational effectiveness and robustness to odometry noise in case of long-term running.

\section{Problem Statement}

Consider a robot equipped with an RGB-D and a noisy odometry sensor in large indoor environment. It moves through the environment along certain trajectory. At each time step $t$, passing through position $p_t$, the robot receives image $I_t$, depth $D_t$, and odometry estimation $\widehat{p_t} = p_t + \epsilon_t$, where $\epsilon_t$ is odometry noise.

The task is to estimate precisely robot's trajectory using noised odometry estimations, images and depths:

$$\tilde{p_t} = A(I_{0..t}, D_{0..t}, \widehat{p}_{0..t});\ \ \ \ E(p_{0..T}, \tilde{p}_{0..T}) \rightarrow \min$$

where $E$ is an error metric.

We use three error metrics in our evaluation: absolute trajectory error $ATE$, relative translation error $E_{trans}$, and relative rotation error $E_{rot}$:

$$ATE(p_{0..T}, \tilde{p}_{0..T}) = \frac{1}{T+1} \sum\limits_{t=0}^T || p_t - \tilde{p}_t ||_2$$

$$E_{trans}(p_{0..T}, \tilde{p}_{0..T}) = \frac{1}{T} \sum\limits_{t=1}^{T} \frac{||  M_{p_{t-1}}^{-1} p_t - M_{\tilde{p}_{t-1}}^{-1} \tilde{p}_t ||_2}{|| p_t - p_{t-1} ||_2}$$

$$E_{rot}(p_{0..T}, \tilde{p}_{0..T}) = \frac{1}{T} \sum\limits_{t=1}^T \frac{\angle (M_{p_{t-1}}^{-1} p_t, M_{\tilde{p}_{t-1}}^{-1} \tilde{p}_t)}{|| p_t - p_{t-1} ||_2}$$

where $M_p$ is the transformation matrix which transforms zero position and orientation into pose $p$; $||x||_2$ is $L_2$-norm of $x$, and $\angle(a, b)$ is the angle between vectors 
$a$ and $b$.

\section{Methods Overview}

\subsection{RTAB-Map}
\label{sec:rtabmap}

\begin{figure}[h]
    \centering
    \includegraphics[width=1.0\textwidth]{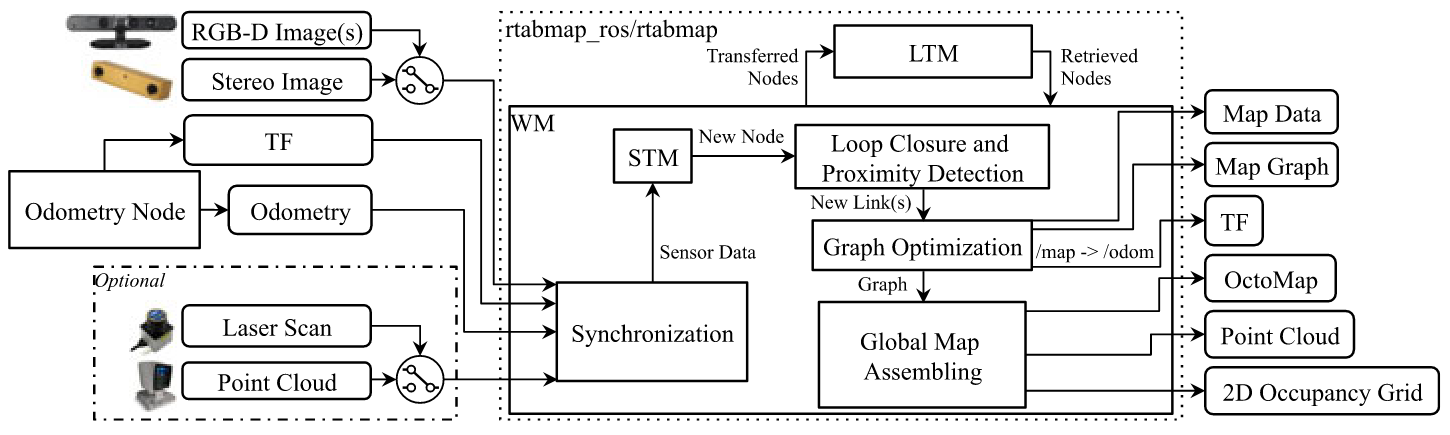}
    \caption{Scheme of RTAB-MAP algorithm (from \cite{labbe2019rtab})}
    \label{fig:rtabmap_scheme}
\end{figure}

RTAB-MAP \cite{labbe2019rtab} is a feature-based SLAM method which takes stereo, RGB-D or laser scan data and builds 2D occupancy grid map and 3D point cloud map. This method is divided by odometry estimation, mapping and loop closure. A scheme of the method is shown in figure \ref{fig:rtabmap_scheme}.

For odometry estimation, RTAB-Map extracts features from images using BRIEF detector \cite{calonder2010brief}. For memory and computational efficiency, RTAB-Map selects key frames from the input flow. New keyframe is added when previous keyframe and current frame has few feature matches (less than certain threshold). Robot motion is tracked by feature matching between current and previous keyframes using PnP RANSAC algorithm \cite{brachmann2017dsac}. Estimated camera position is corrected using Local Bundle Adjustment method \cite{zhang2001incremental} and predictions from previous camera motion.

For mapping, local occupancy grids are used. These grids are received from input depth maps or laser scans. Local maps are fused into global map using a voxel filter. Global map is optimized with loop closures. The loop detection algorithm is based on feature matching on input frames (RGB images or laser scans). Features needed for loop closing are extracted using SURF algorithm \cite{bay2006surf}. The frames are stored as sets of features organized into kd-trees. For effective loop closure, keyframes are organized as a graph where links are constraints taken from keyframe neighborhood or loop detection. In the loop closure stage, this graph is optimized using g2o graph optimization technique \cite{kummerle2011g}.

\begin{figure}[h]
    \centering
    \includegraphics[width=0.7\textwidth]{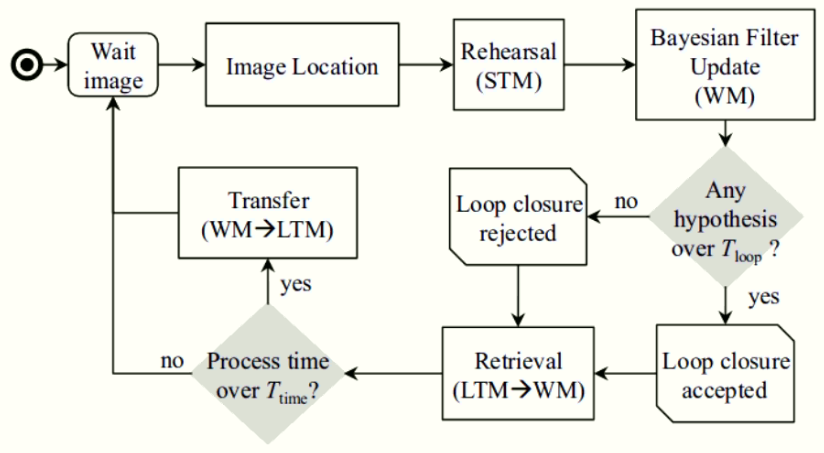}
    \caption{Scheme of RTAB-MAP loop closure methodology (from \cite{labbe2019rtab})}
    \label{fig:rtabmap_loopclosure}
\end{figure}

To store the graph of keyframes, RTAB-MAP uses three levels of memory: working memory (WM), short-term memory (STM), and long-term memory (LTM). WM contains most useful frames, STM contains a sequence of last frames, and LTM contains all frames. Frames that have maximum number of similar features are moved from STM to WM. For loop closure, the frame which is most similar to current is used. The loop closure scheme is shown in fig. \ref{fig:rtabmap_loopclosure}.

\subsection{Voxgraph}

\begin{figure}
    \centering
    \includegraphics[width=0.9\textwidth]{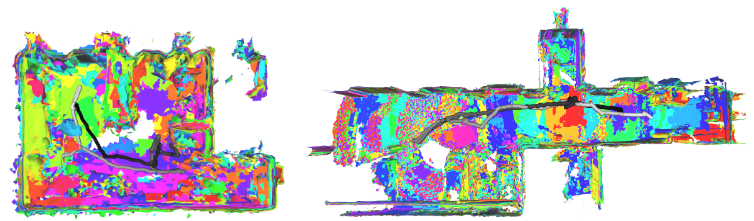}
    \caption{An example of Voxgraph map building (from \cite{reijgwart2019voxgraph}). Different colors represent different submaps}
    \label{fig:voxgraph_submaps}
\end{figure}

Voxgraph \cite{reijgwart2019voxgraph} is a SLAM method which takes odometry and point cloud data as an input and builds pose graph and global map in Truncated Signed Distance Field (TSDF) format. A global map is building as a set of overlapping submaps. These submaps are building as TSDFs by Voxblox \cite{oleynikova2017voxblox} method. New submap is generated with certain period of time, e.g. 20 seconds. An example of submaps and resulting map is shown in fig. \ref{fig:voxgraph_submaps}.

Voxgraph algorithm fuses submaps using submap pose graph. The nodes of the pose graph are sensor positions at start of each submap. The edges of the pose graph are constraints for optimization. They are divided into three kinds: odometry, loop closure, and registration constraints. An odometry constraint between neighbour submaps attaches transformation between these submaps to odometry change from start of the first submap to start of the second one. A loop closure constraint comes from loop closure hypothesis from an external loop closure source. It binds corresponding submaps. A registration constraint binds together a pair of overlapping submaps. Voxgraph algorithm optimizes pose graph with these constraints using non-linear least squares minimization technique.


Such graph-baseed approach lets Voxgraph be robust to odometry noise and significantly reduces memory consumption. According to \cite{schmid2021unified}, a robot is able to operate in large area for 10-15 min with severe odometry drift without collisions, keeping low CPU and memory load. With efficient external loop closure source like DBoW2 \cite{galvez2012bags}, Voxgraph is able to long-term operation without error accumulation and significant memory growth.

\section{Evaluation}
\label{sec:experiments}

\subsection{Experimental setup}

We carried out our experiments in large scene in Habitat simulator. The scene represented one store of university building, with a 100m-long branched corridor, a wide hall and one classroom of size 6x15m. Virtual agent navigated along a manually set route and was tasked to estimate its trajectory and build a map of the environment. The route was set as a sequence of waypoints, and navigation performed using Habitat's built-in ShortestPathFollower algorithm. We used two routes - the former had length 300m and consisted of 12 points, the latter had length 1 km and consisted of 25 points. Linear and angular speed of agent were set to 1.5 m/s and 90 degrees/s respectively.

To model real odometry sensors, we noised ground-truth position with Gaussian noise. We used four degrees of noise - small (linear std 0.0015, angular std 0.003), medium (linear std 0.003, angular std 0.0075), large (linear std 0.0075, angular std 0.015), and extra-large (linear std 0.015, angular std 0.025). Linear noise was added to linear speed, angular noise was added to agent's angle. The noise updated every 10 seconds.

As an RGB-D sensor data, we used ground truth RGB and depth images from the simulator. Depth range was limited by 8m. Both RGB and depth had resolution 320x240, and field of view 90 degrees.

For RTAB-MAP, we used only mapping module, which was feed with noised odometry from simulator. We used RTAB-MAP in 3DoF mode, with extraction of 2000 features from each image and enable ray tracing. Height of the obstacle cells was set from 0.2m to 1.5m. For Voxgraph, we set TSDF voxel size to 0.2, submap creation interval to 20 secs, and TSDF truncation distance to 0.8. 

\subsection{Results}

\begin{table}[h]
    \centering
    \begin{tabular}{|l|c|c|c|c|c|c|c|c|c|}
        \hline
         \multicolumn{2}{|c|}{} & \multicolumn{4}{c}{300m trajectory} \vline & \multicolumn{4}{c}{1km trajectory} \vline\\
         \hline
         \multicolumn{2}{|c|}{Noise} & Small & Medium & Large & X-large & Small & Medium & Large & X-large \\
         \hline
        \multirow{6}{*}{RTAB-MAP} &
        $E_{trans}$ & 0.017 & 0.030 & 0.072 & 0.153 & 0.021 & \textbf{0.037} & \textbf{0.094} & 0.187 \\
        & $(\pm std)$ & 0.002 & 0.0004 & 0.001 & 0.007 & 0.0003 & 0.0006 & 0.001 & 0.004 \\
        \cline{2-10}
        & $E_{rot}$ & 0.071 & 0.106 & 0.181 & 0.276 & 0.213 & 0.308 & 0.571 & 1.29 \\
        & $(\pm std)$ & 0.008 & 0.012 & 0.02 & 0.04 & 0.005 & 0.01 & 0.05 & 0.45 \\
        \cline{2-10}
        & ATE & 0.99 & \textbf{1.91} & \textbf{4.07} & \textbf{7.69} & \textbf{0.92} & \textbf{1.75} & \textbf{5.00} & \textbf{9.77} \\
        & $(\pm std)$ & 0.087 & 0.41 & 0.37 & 0.17 & 0.14 & 0.23 & 0.23 & 1.44 \\
        
        \hline
        
        \multirow{6}{*}{Voxgraph} &
        $E_{trans}$ & \textbf{0.015} & \textbf{0.027} & \textbf{0.065} & \textbf{0.122} & \textbf{0.020} & 0.039 & \textbf{0.094} & \textbf{0.176} \\
        & $(\pm std)$ & 0.0005 & 0.001 & 0.005 & 0.003 & 0.0003 & 0.0006 & 0.001 & 0.004 \\
        \cline{2-10}
        & $E_{rot}$ & \textbf{0.033} & \textbf{0.073} & \textbf{0.152} & \textbf{0.229} & \textbf{0.031} & \textbf{0.071} & \textbf{0.181} & \textbf{0.316} \\
        & $(\pm std)$ & 0.002 & 0.004 & 0.013 & 0.021 & 0.001 & 0.003 & 0.033 & 0.029 \\
        \cline{2-10}
        & ATE & \textbf{0.93} & 2.21 & 5.13 & 10.1 & 2.43 & 5.92 & 10.6 & 20.0 \\
        & $(\pm std)$ & 0.05 & 0.15 & 0.32 & 0.26 & 0.04 & 0.06 & 1.87 & 0.97 \\
        
        \hline
    \end{tabular}
    \caption{Evaluation of RTAB-Map and Voxgraph algorithms on 300m and 1km trajectories: average metric values and standard deviation of 5 runs. $E_{trans}$ is the relative translational error, $E_{rot}$ is the relative rotational error (in degrees per meter), ATE is the absolute trajectory error. For all the metrics, lower is better.}
    \label{tab:results}
\end{table}

\begin{figure}[h]
    \centering
    \includegraphics[width=0.9\textwidth]{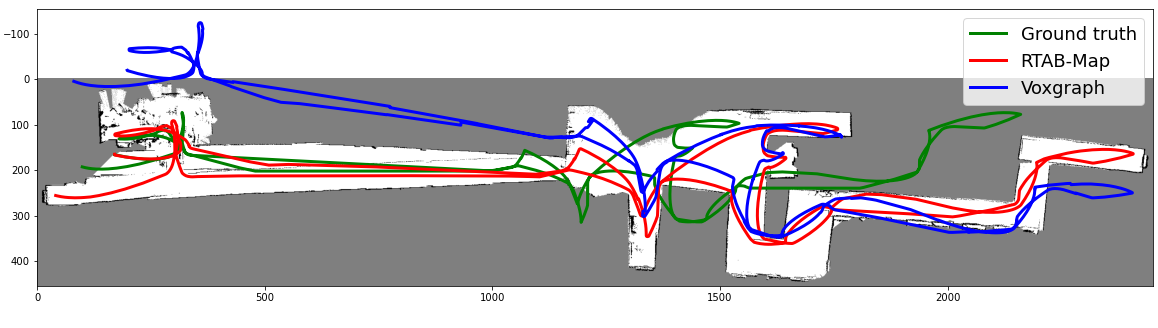}
    \caption{Trajectories estimated by RTAB-Map (red) and Voxgraph (blue) compared to ground-truth trajectory (green). The background is the map built by RTAB-Map, it has a fake corridor in the left part.}
    \label{fig:slam_paths}
\end{figure}

\begin{figure}[h]
    \centering
    \includegraphics[width=0.8\textwidth]{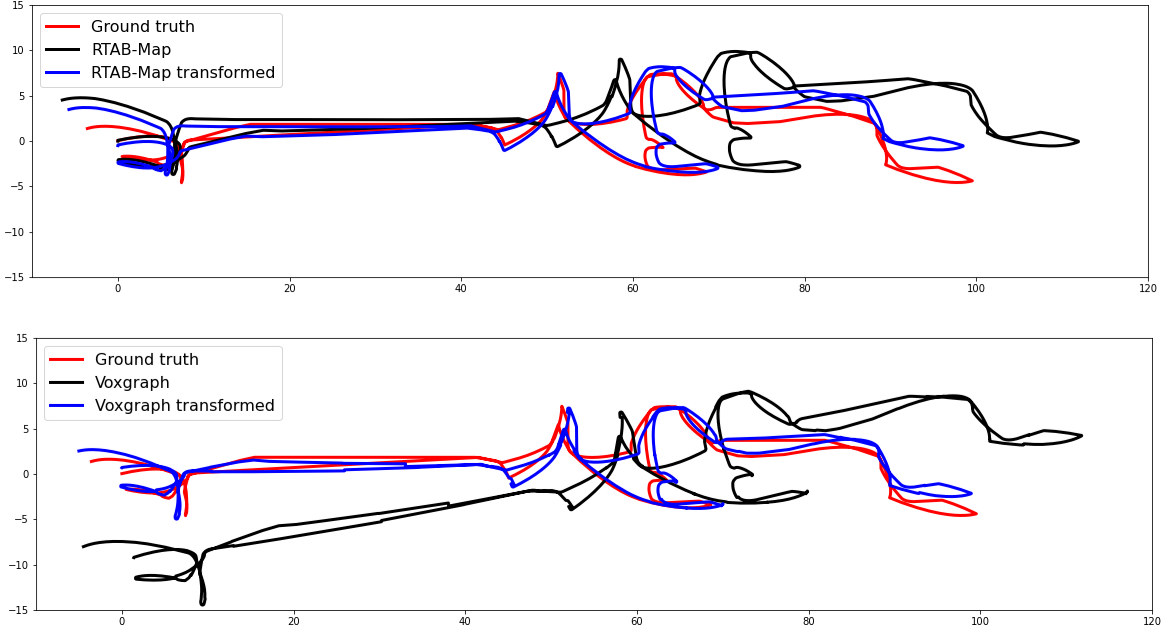}
    \caption{Comparison of RTAB-Map's (top) and Voxgraph's (bottom) estimated trajectory. Red line denotes ground truth trajectory, black line denotes estimated trajectory, and blue line denotes estimated trajectory fitted into ground truth by a transform (translation, rotation, scaling). Voxgraph's trajectory was fitted more accurately, despite it has bigger absolute deviation.}
    \label{fig:path_transforms}
\end{figure}

\begin{table}[h]
    \centering
    \begin{tabular}{|l|c|c|c|c|c|c|c|c|}
        \hline
         & \multicolumn{4}{c}{300m trajectory} \vline & \multicolumn{4}{c}{1km trajectory} \vline\\
         \hline
         Noise & Small & Medium & Large & X-large & Small & Medium & Large & X-large \\
         \hline
        RTAB-MAP & 3.6 & 3.5 & 3.6 & 3.7 & 6.3 & 6.2 & 8.5 & 8.6 \\
        Voxgraph & 0.52 & 0.53 & 0.54 & 0.57 & 1.2 & 1.3 & 1.3 & 1.8 \\
        \hline
    \end{tabular}
    \caption{Memory consumption (in GB) of RTAB-Map and Voxgraph algorithms on 300m and 1km trajectories}
    \label{tab:memory_usage}
\end{table}

\begin{figure}[h]
    \centering
    \includegraphics[width=0.9\textwidth]{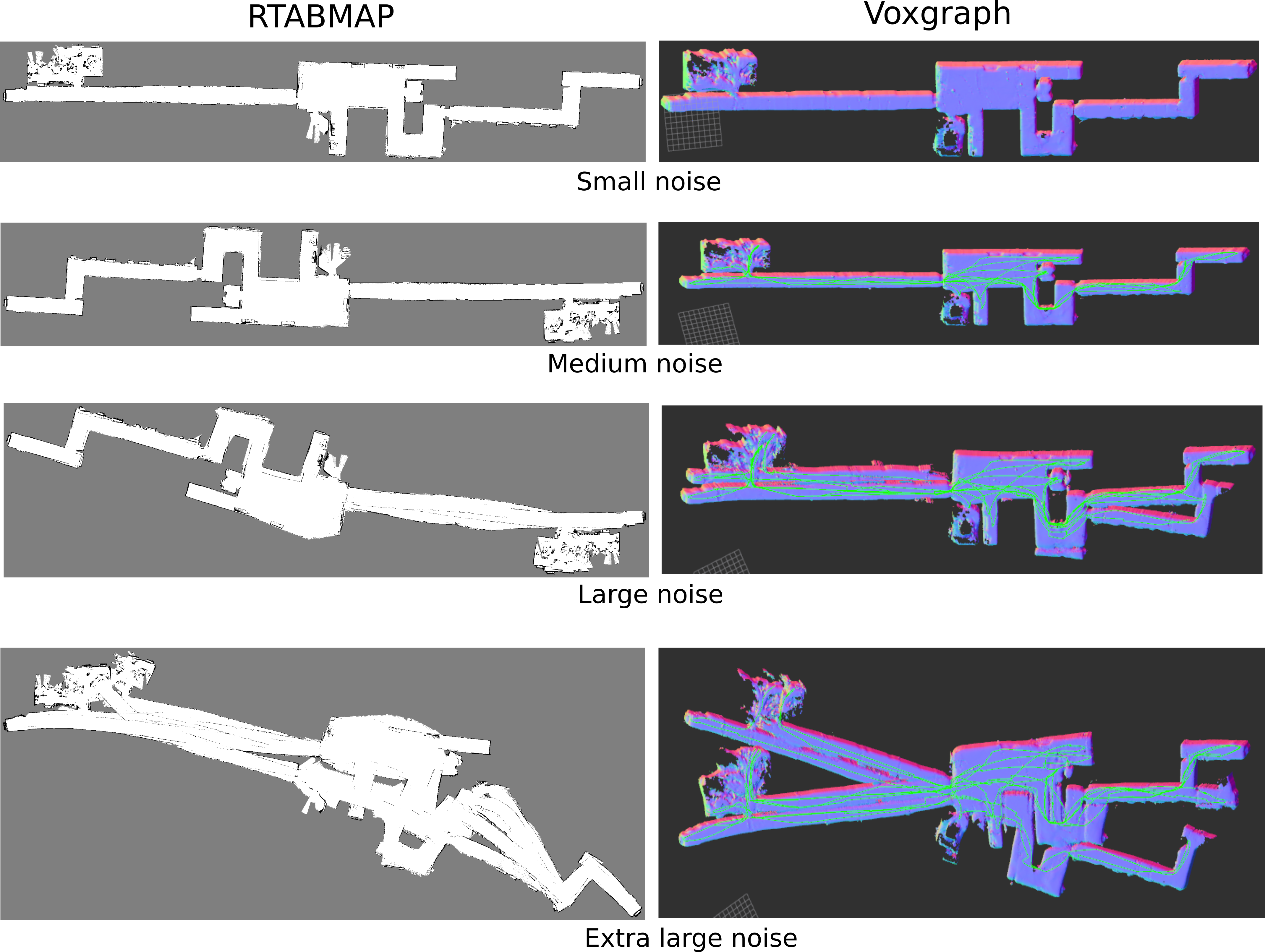}
    \caption{Visual comparison of RTAB-MAP and Voxgraph maps on 1km trajectory.}
    \label{fig:map_comparison}
\end{figure}

\begin{figure}[h]
    \centering
    \includegraphics[width=1.0\textwidth]{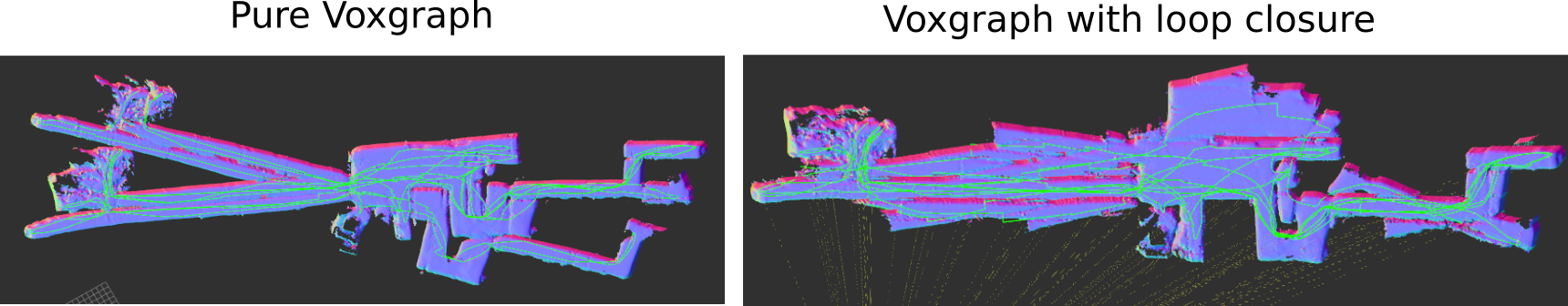}
    \caption{Voxgraph map without loop closure (left) and with DBoW2 loop closure (right).}
    \label{fig:voxgraph_lc_comparison}
\end{figure}

We ran RTAB-Map and Voxgraph methods on 300m and 1km trajectories in Habitat simulator with four degrees of odometry noise: small, medium, large and extra-large. For each degree of noise, we ran both the algorithms 5 times and calculated average metric values and their standard deviation. The resulting metric values are shown in Table \ref{tab:results}. RTAB-Map demonstrated smaller absolute trajectory errors than Voxgraph, however its translation error was comparable to Voxgraph's one, and its rotation error was significantly larger than that of Voxgraph. An example of estimated trajectories (300m, extra-large drift) is shown in fig. \ref{fig:slam_paths}. We can see that Voxgraph's trajectory has bigger absolute deviation than RTAB-Map's one, however it is closer to ground truth relatively (see fig. \ref{fig:path_transforms}). This may be explained by large amount of whole graph optimizations during Voxgraph operation.

Memory consumption values for both algorithms are shown in Table \ref{tab:memory_usage}. As seen in the table, RTAB-Map took up to 8.6 GB of RAM on 1 km trajectory, which can exceed memory limit on some on-board computers. On 300m trajectory, RTAB-Map consumed about 3.6 GB of RAM, which is also a large volume. Voxgraph consumed approximately 5-6 times less memory on both trajectories - up to 0.57 GB on 300m trajectory and up to 1.8 GB on 1 km trajectory with extra-large drift. Such difference in memory volume is explained by difference in the map type (TSDF vs voxel grid), and difference in the graph structure (graph of submaps, vs graph of key frames).

The built maps of both algorithms on 1km trajectory are shown in Fig. \ref{fig:map_comparison}. With small and medium noise degrees, both algorithms built maps close to the ideal, but with large and extra-large noises, both algorithms failed to build a consistent map. RTAB-Map built fake corridors in the middle part of the map, whereas Voxgraph built fake corridors in the left and the right parts of the map. So, RTAB-Map's loop closure started failing in long distances and severe odometry noises. Voxgraph, which has no built-in loop closure, also fails to fuse large map under severe odometry noise. We also tried Voxgraph with DBoW2 loop closure, this helped reduce number of "corridor bifurcation" cases, but did not eliminate all of them (see fig. \ref{fig:voxgraph_lc_comparison}). The issue of interaction of Voxgraph and loop closure methods requires additional research.

\section{Conclusion and Future Work}

In this paper, we considered a problem of simultaneous localization and mapping in large indoor environments. For our study, we chose keyframe-based SLAM method RTAB-Map and submap-based SLAM method Voxgraph, both of them are positioned as suitable for long-term navigation and large environments. We carried out experimental evaluation of them on an indoor simulated setup with 300m and 1km trajectories.

The experiments have shown that the submap-based method Voxgraph overcomes RTAB-Map in terms of memory consumption and relative errors, but loses in absolute trajectory error. On long trajectories, RTAB-Map's memory consumption may exceed RAM limit of a computer of a mobile robotic system. Also, in case of long trajectory and high degree of odometry noise, both methods fail to build plausible map.

Our empirical evaluation has shown that long-term navigation in large environments is a challenging problem that is lacking an universal solution and needs additional research. The perspective fields of this research are memory-efficient loop closure methods, and topological approaches to SLAM.




%
%
%
\bibliographystyle{splncs03_unsrt}
\bibliography{mybibliography}

\begin{thebibliography}{10}
\providecommand{\url}[1]{\texttt{#1}}
\providecommand{\urlprefix}{URL }

\bibitem{gonzalez2017supervisory}
Gonzalez, A.G., Alves, M.V., Viana, G.S., Carvalho, L.K., Basilio, J.C.:
  Supervisory control-based navigation architecture: a new framework for
  autonomous robots in industry 4.0 environments. IEEE Transactions on
  Industrial Informatics  14(4),  1732--1743 (2017)

\bibitem{papachristos2019autonomous}
Papachristos, C., Khattak, S., Mascarich, F., Alexis, K.: Autonomous navigation
  and mapping in underground mines using aerial robots. In: 2019 IEEE Aerospace
  Conference. pp. 1--8. IEEE (2019)

\bibitem{tang2019topological}
Tang, L., Wang, Y., Ding, X., Yin, H., Xiong, R., Huang, S.: Topological
  local-metric framework for mobile robots navigation: a long term perspective.
  Autonomous Robots  43(1),  197--211 (2019)

\bibitem{choi2020development}
Choi, J., Park, J., Jung, J., Lee, Y., Choi, H.T.: Development of an autonomous
  surface vehicle and performance evaluation of autonomous navigation
  technologies. International Journal of Control, Automation and Systems
  18(3),  535--545 (2020)

\bibitem{mur2015orb}
Mur-Artal, R., Montiel, J.M.M., Tardos, J.D.: Orb-slam: a versatile and
  accurate monocular slam system. IEEE transactions on robotics  31(5),
  1147--1163 (2015)

\bibitem{engel2014lsd}
Engel, J., Sch{\"o}ps, T., Cremers, D.: Lsd-slam: Large-scale direct monocular
  slam. In: European conference on computer vision. pp. 834--849. Springer
  (2014)

\bibitem{teed2021droid}
Teed, Z., Deng, J.: Droid-slam: Deep visual slam for monocular, stereo, and
  rgb-d cameras. Advances in Neural Information Processing Systems  34,
  16558--16569 (2021)

\bibitem{labbe2019rtab}
Labb{\'e}, M., Michaud, F.: Rtab-map as an open-source lidar and visual
  simultaneous localization and mapping library for large-scale and long-term
  online operation. Journal of Field Robotics  36(2),  416--446 (2019)

\bibitem{reijgwart2019voxgraph}
Reijgwart, V., Millane, A., Oleynikova, H., Siegwart, R., Cadena, C., Nieto,
  J.: Voxgraph: Globally consistent, volumetric mapping using signed distance
  function submaps. IEEE Robotics and Automation Letters  5(1),  227--234
  (2019)

\bibitem{savva2019habitat}
Savva, M., Kadian, A., Maksymets, O., Zhao, Y., Wijmans, E., Jain, B., Straub,
  J., Liu, J., Koltun, V., Malik, J., et~al.: Habitat: A platform for embodied
  ai research. In: Proceedings of the IEEE/CVF International Conference on
  Computer Vision. pp. 9339--9347 (2019)

\bibitem{smith1988stochastic}
Smith, R., Self, M., Cheeseman, P.: A stochastic map for uncertain spatial
  relationships. In: Proceedings of the 4th international symposium on Robotics
  Research. pp. 467--474 (1988)

\bibitem{klein2007parallel}
Klein, G., Murray, D.: Parallel tracking and mapping for small ar workspaces.
  In: 2007 6th IEEE and ACM international symposium on mixed and augmented
  reality. pp. 225--234. IEEE (2007)

\bibitem{triggs1999bundle}
Triggs, B., McLauchlan, P.F., Hartley, R.I., Fitzgibbon, A.W.: Bundle
  adjustment—a modern synthesis. In: International workshop on vision
  algorithms. pp. 298--372. Springer (1999)

\bibitem{mur2017orb}
Mur-Artal, R., Tard{\'o}s, J.D.: Orb-slam2: An open-source slam system for
  monocular, stereo, and rgb-d cameras. IEEE transactions on robotics  33(5),
  1255--1262 (2017)

\bibitem{geiger2013vision}
Geiger, A., Lenz, P., Stiller, C., Urtasun, R.: Vision meets robotics: The
  kitti dataset. The International Journal of Robotics Research  32(11),
  1231--1237 (2013)

\bibitem{min2021voldor+}
Min, Z., Dunn, E.: Voldor+ slam: For the times when feature-based or direct
  methods are not good enough. In: 2021 IEEE International Conference on
  Robotics and Automation (ICRA). pp. 13813--13819. IEEE (2021)

\bibitem{gomez2020hybrid}
Gomez, C., Fehr, M., Millane, A., Hernandez, A.C., Nieto, J., Barber, R.,
  Siegwart, R.: Hybrid topological and 3d dense mapping through autonomous
  exploration for large indoor environments. In: 2020 IEEE International
  Conference on Robotics and Automation (ICRA). pp. 9673--9679. IEEE (2020)

\bibitem{oleynikova2017voxblox}
Oleynikova, H., Taylor, Z., Fehr, M., Siegwart, R., Nieto, J.: Voxblox:
  Incremental 3d euclidean signed distance fields for on-board mav planning.
  In: 2017 IEEE/RSJ International Conference on Intelligent Robots and Systems
  (IROS). pp. 1366--1373. IEEE (2017)

\bibitem{besl1992method}
Besl, P.J., McKay, N.D.: Method for registration of 3-d shapes. In: Sensor
  fusion IV: control paradigms and data structures. vol. 1611, pp. 586--606.
  Spie (1992)

\bibitem{schmid2021unified}
Schmid, L., Reijgwart, V., Ott, L., Nieto, J., Siegwart, R., Cadena, C.: A
  unified approach for autonomous volumetric exploration of large scale
  environments under severe odometry drift. IEEE Robotics and Automation
  Letters  6(3),  4504--4511 (2021)

\bibitem{niijima2020city}
Niijima, S., Umeyama, R., Sasaki, Y., Mizoguchi, H.: City-scale
  grid-topological hybrid maps for autonomous mobile robot navigation in urban
  area. In: 2020 IEEE/RSJ International Conference on Intelligent Robots and
  Systems (IROS). pp. 2065--2071. IEEE (2020)

\bibitem{schmuck2016hybrid}
Schmuck, P., Scherer, S.A., Zell, A.: Hybrid metric-topological 3d occupancy
  grid maps for large-scale mapping. IFAC-PapersOnLine  49(15),  230--235
  (2016)

\bibitem{sturm12iros}
Sturm, J., Engelhard, N., Endres, F., Burgard, W., Cremers, D.: A benchmark for
  the evaluation of rgb-d slam systems. In: Proc. of the International
  Conference on Intelligent Robot Systems (IROS) (Oct 2012)

\bibitem{burri2016euroc}
Burri, M., Nikolic, J., Gohl, P., Schneider, T., Rehder, J., Omari, S.,
  Achtelik, M.W., Siegwart, R.: The euroc micro aerial vehicle datasets. The
  International Journal of Robotics Research  35(10),  1157--1163 (2016)

\bibitem{bokovoy2021assessment}
Bokovoy, A., Muraviev, K.: Assessment of map construction in vslam. In: 2021
  International Siberian Conference on Control and Communications (SIBCON). pp.
  1--6. IEEE (2021)

\bibitem{mingachev2021comparative}
Mingachev, E., Lavrenov, R., Magid, E., Svinin, M.: Comparative analysis of
  monocular slam algorithms using tum and euroc benchmarks. In: Proceedings of
  15th International Conference on Electromechanics and Robotics" Zavalishin's
  Readings". pp. 343--355. Springer (2021)

\bibitem{zhang2021survey}
Zhang, S., Zheng, L., Tao, W.: Survey and evaluation of rgb-d slam. IEEE Access
   9,  21367--21387 (2021)

\bibitem{shi2020we}
Shi, X., Li, D., Zhao, P., Tian, Q., Tian, Y., Long, Q., Zhu, C., Song, J.,
  Qiao, F., Song, L., et~al.: Are we ready for service robots? the
  openloris-scene datasets for lifelong slam. In: 2020 IEEE international
  conference on robotics and automation (ICRA). pp. 3139--3145. IEEE (2020)

\bibitem{hong2021radar}
Hong, Z., Petillot, Y., Wallace, A., Wang, S.: Radar slam: A robust slam system
  for all weather conditions. arXiv preprint arXiv:2104.05347  (2021)

\bibitem{calonder2010brief}
Calonder, M., Lepetit, V., Strecha, C., Fua, P.: Brief: Binary robust
  independent elementary features. In: European conference on computer vision.
  pp. 778--792. Springer (2010)

\bibitem{brachmann2017dsac}
Brachmann, E., Krull, A., Nowozin, S., Shotton, J., Michel, F., Gumhold, S.,
  Rother, C.: Dsac-differentiable ransac for camera localization. In:
  Proceedings of the IEEE conference on computer vision and pattern
  recognition. pp. 6684--6692 (2017)

\bibitem{zhang2001incremental}
Zhang, Z., Shan, Y.: Incremental motion estimation through local bundle
  adjustment  (2001)

\bibitem{bay2006surf}
Bay, H., Tuytelaars, T., Gool, L.V.: Surf: Speeded up robust features. In:
  European conference on computer vision. pp. 404--417. Springer (2006)

\bibitem{kummerle2011g}
K{\"u}mmerle, R., Grisetti, G., Strasdat, H., Konolige, K., Burgard, W.: g 2 o:
  A general framework for graph optimization. In: 2011 IEEE International
  Conference on Robotics and Automation. pp. 3607--3613. IEEE (2011)

\bibitem{galvez2012bags}
G{\'a}lvez-L{\'o}pez, D., Tardos, J.D.: Bags of binary words for fast place
  recognition in image sequences. IEEE Transactions on Robotics  28(5),
  1188--1197 (2012)

\end{thebibliography}

\end{document}